\documentclass[11pt]{article}

\usepackage[utf8]{inputenc}
\usepackage[T1]{fontenc}
\usepackage{lmodern}
\usepackage[margin=1in]{geometry}
\usepackage{amsmath,amssymb,amsthm}
\usepackage{graphicx}
\usepackage{booktabs}
\usepackage{hyperref}
\usepackage{xcolor}
\usepackage{natbib}
\usepackage{float}
\usepackage{enumitem}
\usepackage{url}
\usepackage{microtype}

\hypersetup{colorlinks=true, linkcolor=blue!70!black, citecolor=blue!70!black, urlcolor=blue!70!black}

\newtheorem{proposition}{Proposition}
\newcommand{\sstar}{s^\star}
\newcommand{\Xp}{X_P}

\title{Filtered ANN as a Phase Transition:\\
When Selectivity-Estimation Error Causes Plan Regret}

\author{
  Madhulatha Mandarapu\thanks{madhulatha@samyama.ai} \and
  Sandeep Kunkunuru\thanks{sandeep@samyama.ai}
}
\date{%
  VaidhyaMegha Private Limited, India\\[2pt]
  \url{https://samyama.ai/}\\[8pt]
  June 2026
}

\begin{document}
\maketitle

\begin{abstract}
A filtered approximate-nearest-neighbor (ANN) query returns the $k$ nearest vectors \emph{among those
satisfying an attribute predicate} $P$ of selectivity $s$. The best execution strategy---pre-filter,
post-filter, or in-filter---changes with $s$, so a system must \emph{estimate} $s$ and choose. We model
this as an $\arg\max$ over a landscape with \emph{phases} (regions where each strategy wins)
separated by \emph{boundaries}, and show that selectivity-estimation error produces plan
\emph{regret}---recall lost versus the oracle strategy---\emph{only in the critical regions around
those boundaries}. The regret is a wedge of log-width equal to the multiplicative estimation error
$\varepsilon$ and height equal to the local cliff $|V'(\sstar)|\,\varepsilon$; the flip-margin
$1/|V'(\sstar)|$ is the condition number of a sibling cardinality-estimation study reappearing as the
local boundary theory. The two phase boundaries follow from independent mathematics: \emph{order
statistics} place the post-filter cliff at $s\approx k/K$, and \emph{site percolation} places the
in-filter cliff at $s_c\approx 0.83/M$ for graph degree $M$ (corpus-size independent). Criticality
exists only under a \emph{constrained budget} $B<\sqrt{kn}$. Under pre-registered decision rules we
confirm, on synthetic sweeps and real SIFT1M, that regret concentrates $\sim\!290\times$ at the
boundary and that the regret curves obey a \emph{finite-size scaling collapse} onto one universal wedge
across two decades of corpus size. A real approximate index does not mis-locate the boundary, but a
biased cost model opens a persistent miscalibration band that estimation-error robustness cannot fix.
The contribution is a characterization, not a new index. Code and the full pre-registration are public.\footnote{\url{https://github.com/samyama-ai/filtered-ann-regret}}
\end{abstract}

\section{Introduction}\label{sec:intro}
Production vector search rarely runs unconstrained: a query asks for the $k$ nearest vectors
\emph{among those satisfying a predicate} $P$ (e.g.\ \texttt{category='shoes' AND price<100}). The
predicate's \emph{selectivity} $s=|\Xp|/n$ (the fraction of the $n$-vector corpus passing $P$) spans
near-$0$ to near-$1$, and the right execution strategy moves with it
\citep{patel2024acorn,gollapudi2023filtered,simhadri2024bigann}---bad estimates yield bad plans, the
vector-search echo of relational query optimization \citep{leis2015job}:
\emph{pre-filter} (materialize $\Xp$, then search within it) wins at low $s$;
\emph{post-filter} (run unfiltered ANN, drop violators) wins at high $s$;
and an \emph{in-filter} traversal that respects $P$ during search targets the middle.
A deployed system must \emph{estimate} $s$ and pick a strategy---exactly the access-path-selection
problem of \citet{selinger1979access}, now over a recall/latency objective. Vespa compares an estimated
hit-ratio against two thresholds \citep{vespa2023constrained}; AlloyDB switches strategy mid-query on a
bad estimate; and a recent learned planner predicts the plan from predicate statistics
\citep{ganwang2026planning}. Strategy selection from an estimate is, in short, established practice.

What is \emph{not} established is a quantitative account of \emph{when, how badly, and how universally a
wrong estimate hurts}. We provide one, by recognizing the structure underneath: \textbf{filtered-ANN
strategy selection is a phase-transition system.} Selectivity is an order parameter; the strategies are
phases; and---this is the organizing observation---estimation error causes regret \emph{only} in the
critical regions around the phase boundaries. Deep inside a phase the best strategy dominates by a wide
margin, so even a badly wrong estimate picks correctly (regret $\approx 0$); near a boundary the two
strategies are nearly tied, so a small error flips the choice and pays the local cliff. \emph{Robustness
is criticality.}

\paragraph{Contributions.}
\begin{enumerate}[leftmargin=*,topsep=2pt,itemsep=1pt]
\item A \textbf{phase-diagram model} of filtered-ANN strategy selection whose two boundaries have
closed-form locations from independent mathematics: the post-filter cliff at $s\approx k/K$ from
\emph{order statistics} (\S\ref{sec:phases}), and the in-filter cliff at $s_c\approx 0.83/M$ from
\emph{site percolation} (\S\ref{sec:phases},\,\S\ref{sec:exp})---empirically corpus-size-independent,
refining the folklore $\log n/M$.
\item A \textbf{criticality law} (\S\ref{sec:crit}): selection regret is a wedge of log-width
$\varepsilon$ (the multiplicative estimation error) and height $|V'(\sstar)|\,\varepsilon$; the
flip-margin $1/|V'(\sstar)|$ is a condition number \citep{mandarapu2026regimes}. A \emph{constrained-budget}
law $B<\sqrt{kn}$ delimits when any of this matters.
\item A \textbf{finite-size scaling collapse} (\S\ref{sec:exp}): across two decades of corpus size and
all error magnitudes, regret curves collapse onto one universal wedge---selection regret is
scale-invariant, to our knowledge a new observation for vector search.
\item An honest \textbf{model-mismatch} result (\S\ref{sec:exp}): a real approximate index does not
mis-locate the boundary (the criticality picture holds on real ANN), but a biased cost model opens a
\emph{persistent} miscalibration band that estimation-error robustness cannot remove.
\end{enumerate}
Everything is validated on synthetic sweeps and real SIFT1M under \textbf{pre-registered} decision
rules. We claim a \emph{characterization}, not a new index or planner; \S\ref{sec:related} credits the
substantial prior art this builds on.

\section{Problem and Model}\label{sec:model}
A query is $(q,P,k)$: return the $k$ vectors nearest $q$ among $\Xp=\{x:P(x)\}$, with
$s=|\Xp|/n$. We measure quality by \textbf{recall@$k$ at a fixed compute budget} $B$ (distance
evaluations, hardware-independent), and define \emph{strategy-selection regret} as the recall@$k$ lost
by the chosen strategy relative to the oracle-best strategy under the true $s$. Write $M(a,s)$ for the
recall of strategy $a\in\{\text{pre},\text{post},\text{in}\}$ at selectivity $s$. A planner sees an
estimate $\hat s=s\cdot Q$ (a multiplicative error; $\varepsilon=|\ln Q|$ is the log error, the
$q$-error of \citet{moerkotte2009qerror}) and picks $a_{\hat s}=\arg\max_a M(a,\hat s)$; the oracle uses
the true $s$. The regret is $\Delta R = M(a^\star(s),s)-M(a_{\hat s}(\hat s),s)\ge 0$.

\paragraph{Strategy objectives under budget $B$.}
Pre-filter materializes $\Xp$ (cost $\propto sn$): exact when $sn\le B$, else a random $B$-subset is
scanned, giving $M(\text{pre},s)=\min(1,B/(sn))$, \emph{decreasing} in $s$ and \emph{indifferent} to any
correlation between $P$ and $q$. Post-filter examines the unfiltered top-$K$ (we identify the budget with
the candidate-list size, $K=B$) and keeps the eligible ones, giving $M(\text{post},s)$ \emph{increasing}
in $s$ (\S\ref{sec:phases}). The value gap $V(s)=M(\text{pre},s)-M(\text{post},s)$ changes sign at a
boundary $\sstar$.

\section{The Phase Diagram}\label{sec:phases}
\paragraph{Post-filter cliff (order statistics).}
Rank the corpus by distance to $q$; the unfiltered top-$K$ is a prefix of that ranking, and the true
filtered top-$k$ are the first $k$ eligible points. Under an uncorrelated predicate each ranked point is
eligible i.i.d.\ with probability $s$, so the eligible count in the top-$K$ is $E\sim\mathrm{Binomial}(K,s)$,
and because a prefix recovers the \emph{first} $\min(k,E)$ eligible points exactly,
\begin{equation}\label{eq:post}
M(\text{post},s)=\frac{\mathbb{E}[\min(k,\mathrm{Binomial}(K,s))]}{k},
\end{equation}
a sigmoid in $s$ with knee at $s\approx k/K$ (mean eligible $Ks=k$). Eq.~\eqref{eq:post} is exact for the
uncorrelated model, independent of geometry; it is our reproduction gate.

\paragraph{In-filter cliff (percolation).}
A graph index (HNSW/Vamana \citep{malkov2020hnsw,subramanya2019diskann}) is navigable as a whole, but
restricting greedy search to the eligible nodes $\Xp$ is \emph{site percolation}: keeping an $s$-fraction
of a degree-$M$ navigable graph, a giant connected component---hence a reachable target---survives only
above a threshold $s_c$. The connectivity folklore reads $s_c\sim\log n/M$; we measure the
\emph{giant-component} (navigability) threshold and find $s_c\approx0.83/M$, \emph{independent of} $n$
(\S\ref{sec:exp}). The relevant random-graph theory is that of $k$-nearest-neighbor / geometric graphs
\citep{penrose2003rgg}.

\paragraph{Constrained budget.}
Pre-filter is exact for $s\le B/n$ and post-filter saturates for $s\gtrsim k/B$; these ``good'' regions
overlap iff $B\ge\sqrt{kn}$. So a second transition, in the \emph{budget}, separates a free-lunch regime
($B\ge\sqrt{kn}$: both strategies near-perfect over a band, estimation error harmless) from a
\emph{contested} regime ($B<\sqrt{kn}$: a gap where neither end-strategy is good---the literature's
``middle regime''). Criticality, and everything below, lives in the contested regime.

\section{Regret Is Critical}\label{sec:crit}
A threshold planner picks pre iff $\hat s<\sstar$, so it mis-picks iff $s$ and $\hat s=sQ$ straddle
$\sstar$, i.e.\ iff $|\ln s-\ln\sstar|<\varepsilon$ on the appropriate side. Hence:

\begin{proposition}[Flip-margin law]\label{prop:flip}
To first order near a boundary $\sstar$, the mis-pick (flip) region has log-width equal to the
estimation error $\varepsilon$, and the regret inside it is $\Delta R(s)\approx|V'(\sstar)|\,|\ln
s-\ln\sstar|$, rising to a peak $\approx|V'(\sstar)|\,\varepsilon$. The half-width of the band in which a
unit log-error is dangerous is the \emph{flip-margin} $1/|V'(\sstar)|$.
\end{proposition}

\noindent Width is set by the estimation error; \emph{height} by the boundary sharpness $|V'(\sstar)|$.
The flip-margin is exactly the per-decision condition number of the sibling cardinality-estimation study
\citep{mandarapu2026regimes}: its smooth-argmin condition number reappears here as the
\emph{local} theory of each phase boundary, while percolation and order statistics supply the
\emph{global} phase structure---an average-case companion to worst-case robust query processing
\citep{haritsa2020robust}. Two consequences we test: (i) interiors are safe ($\Delta R\approx0$);
(ii) rescaling $x=(\ln s-\ln\sstar)/\varepsilon$ and $y=\Delta R/(|V'(\sstar)|\,\varepsilon)$ should
collapse all curves---across error magnitudes and corpus sizes---onto one universal wedge.

\section{Experiments}\label{sec:exp}
\paragraph{Setup.} All hypotheses, thresholds, and decision rules were \textbf{frozen before any
results} (a dated, pre-data amendment sharpened the flip-margin prediction). The controlled $n$-sweep
uses clustered-Gaussian corpora (only $n$ varies); \textbf{SIFT1M} \citep{jegou2011sift} is the
real-geometry anchor. Post-filter recall is averaged over random predicate draws (a single draw on
clustered data is overdispersed versus Eq.~\eqref{eq:post}); budgets are in the contested regime
$B<\sqrt{kn}$; $k=10$; statistics use $\ge\!1000$ query samples per cell. Code, data scripts, tests
($29$), and the pre-registration are public.

\begin{figure}[H]\centering
\includegraphics[width=\linewidth]{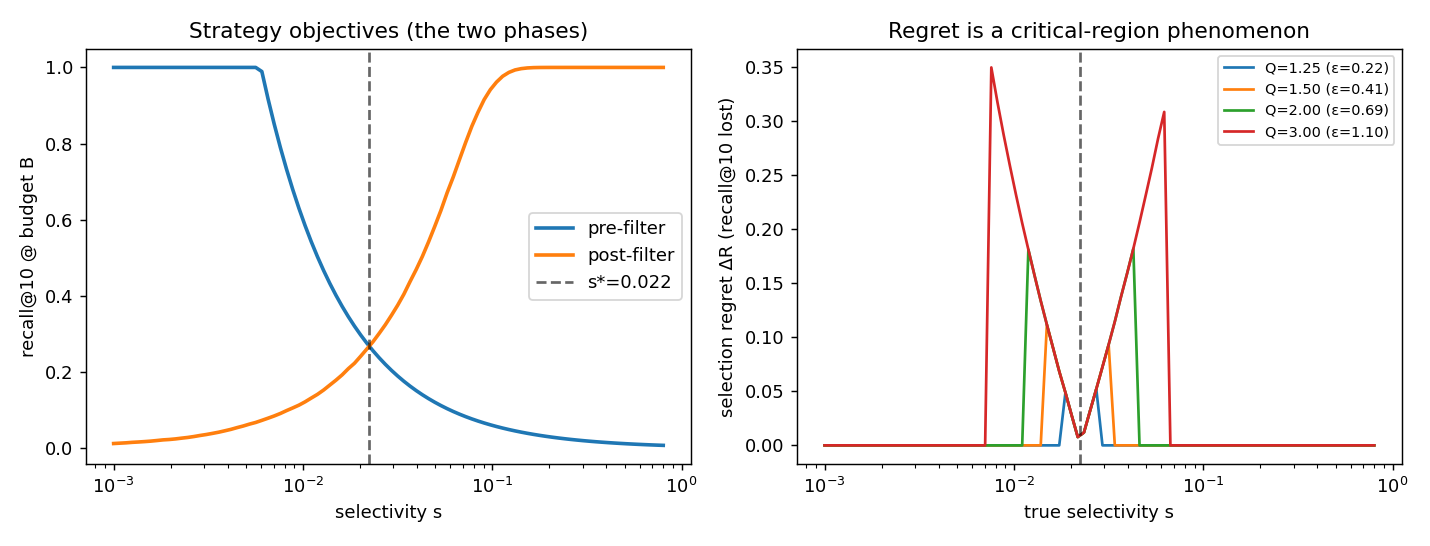}
\caption{The two phases (left) and the regret double-wedge (right): regret concentrates at $\sstar$ and
is $\approx0$ elsewhere; over/under-estimation each open a danger zone on one side. A near-boundary
mis-pick costs up to $\sim0.35$ recall@10.}
\label{fig:crit}
\end{figure}

\paragraph{H0 / H2: criticality.} Eq.~\eqref{eq:post} reproduces on real rankings (gate passed). The
regret is sharply critical (Fig.~\ref{fig:crit}): at $n\in\{10^5,10^6,10^7\}$ \emph{and on real SIFT1M},
interior mean $\Delta R=0.0005$, a $\sim\!290\times$ concentration at the boundary, one-sided flip-width
linear in $\varepsilon$ (slope $0.97$), and peak $\propto|V'(\sstar)|\,\varepsilon$ (Spearman $\rho=1.0$)
---confirming Proposition~\ref{prop:flip}. The crossover scales as $\sstar\!\sim\!1/\sqrt{n}$
($0.010\!\to\!0.0032\!\to\!0.0010$); SIFT1M lands on the synthetic $n=10^6$ point.

\begin{figure}[H]\centering
\includegraphics[width=\linewidth]{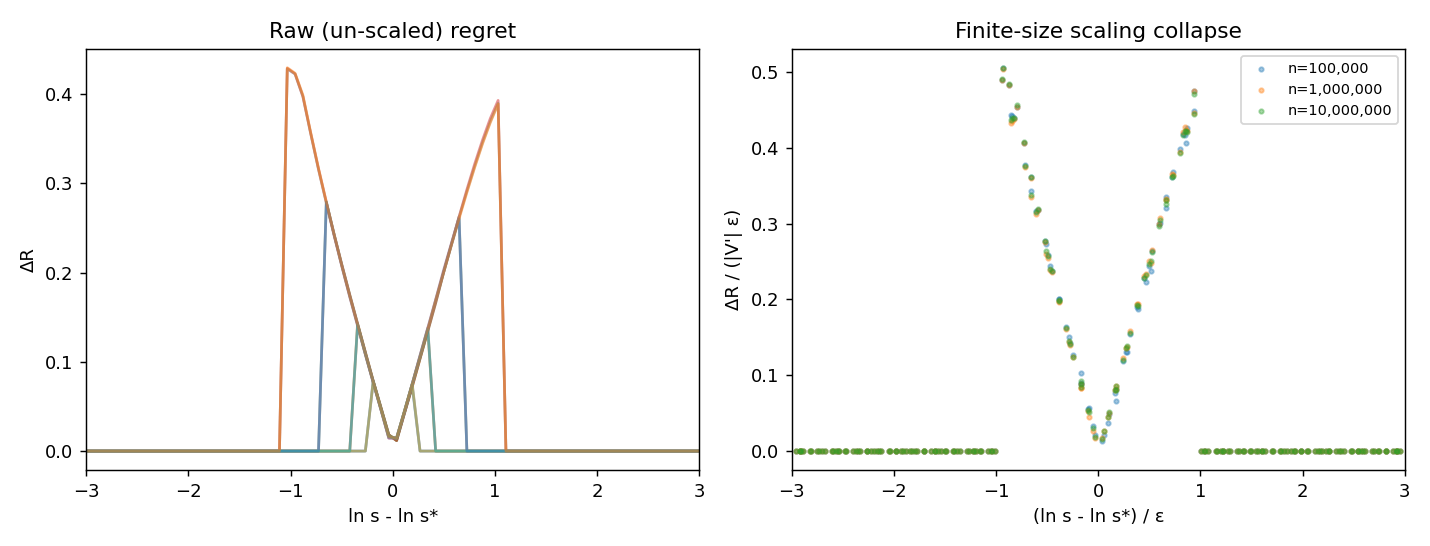}
\caption{Finite-size scaling collapse. Left: raw regret wedges differ in width/height across error
magnitude and $n$. Right: rescaling by $\varepsilon$ and $|V'(\sstar)|\varepsilon$ collapses all of them
onto one universal curve.}
\label{fig:collapse}
\end{figure}

\paragraph{H3: scaling collapse.} Under the Proposition~\ref{prop:flip} rescaling, regret curves from
all $(n,Q)$ cells collapse onto one universal wedge (Fig.~\ref{fig:collapse}): within-bin RMSE $0.015$
($\le0.05$), a $15.7\times$ reduction in inter-curve spread, over two decades of $n$. \emph{Selection
regret is scale-invariant.}

\begin{figure}[H]\centering
\includegraphics[width=\linewidth]{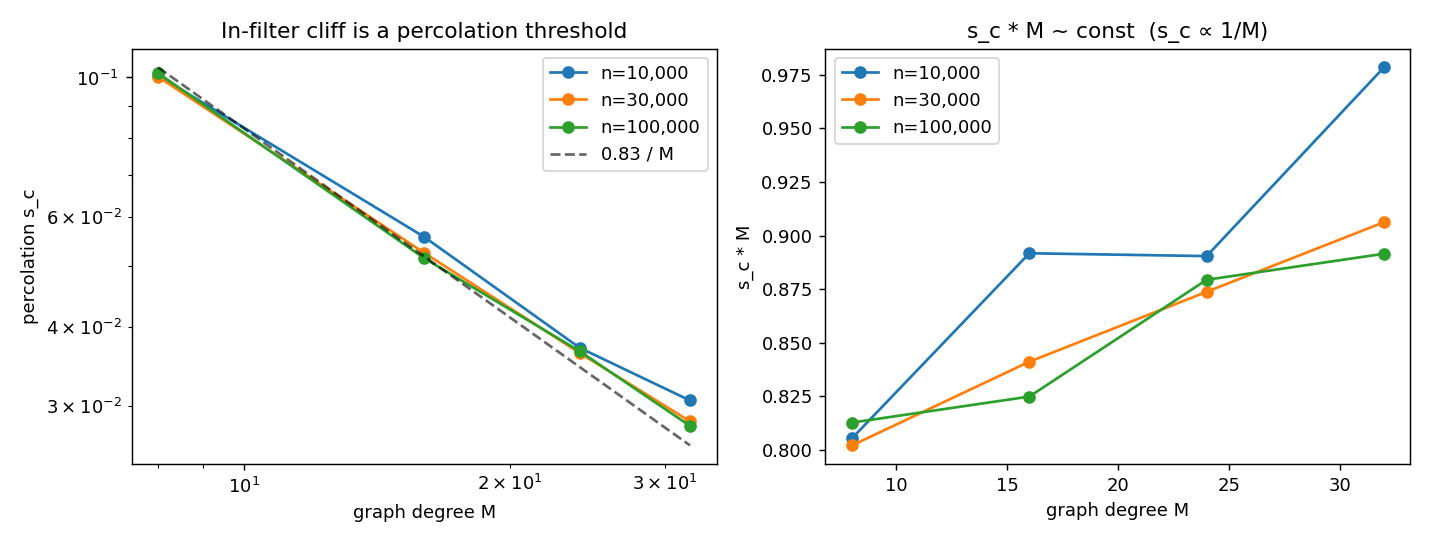}
\caption{The in-filter cliff is a site-percolation transition. Left: $s_c$ vs degree $M$ on log--log,
all corpus sizes on the $0.83/M$ line. Right: $s_c\cdot M$ is flat in $n$.}
\label{fig:perc}
\end{figure}

\paragraph{H1b: percolation.} Building a navigable small-world graph (nearest-neighbor plus long-range
links, so it is one component at $s=1$) and measuring the giant-component fraction of $G[\Xp]$ across
$n\in\{10^4,3\!\cdot\!10^4,10^5\}\times M\in\{8,16,24,32\}$, the in-filter cliff is a sharp percolation
transition at $s_c\approx0.83/M$, \emph{independent of $n$} (Fig.~\ref{fig:perc}; fit slope in $\ln M$ is
$-0.91$, in $\ln\ln n$ is $-0.20$, $R^2=0.998$). This refines the pre-registered $\log n/M$ to the
giant-component law that actually governs navigability. In the phase diagram the in-filter boundary is
thus a horizontal line at $s\approx0.83/M$, while the post cliff ($k/K$) and the crossover ($1/\sqrt n$)
move with scale.

\begin{figure}[H]\centering
\includegraphics[width=\linewidth]{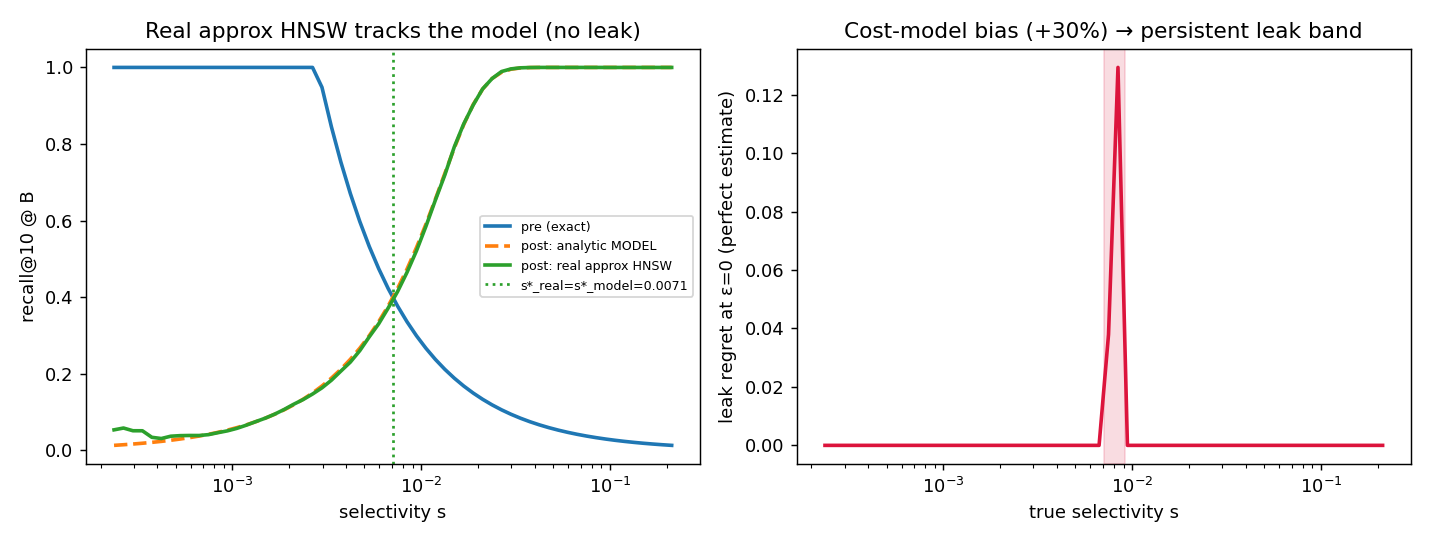}
\caption{Left: a real approximate HNSW post-filter tracks the analytic model; the crossover does not
move (no leak). Right: a $+30\%$ cost-model bias opens a persistent regret band at the boundary, present
even with a perfect per-query estimate.}
\label{fig:leak}
\end{figure}

\paragraph{Model mismatch (beyond the frozen tests).} Is the wedge an artifact of a perfect-model
threshold selector? We test with a \emph{real} approximate HNSW post-filter and with a controlled
cost-model bias $b$ (Fig.~\ref{fig:leak}). The real index \emph{does not} mis-locate the boundary: at
ef$=B$ its induced crossover matches the analytic one to within grid resolution, so the criticality
picture holds on real ANN. But a biased model ($\sstar$ off by $\pm20$--$40\%$) opens a \emph{persistent}
miscalibration band---regret present even at $\varepsilon=0$ (a perfect per-query estimate), peak
$\Delta R\approx0.13$--$0.34$---that robustness to estimation noise cannot fix. There are thus
\emph{two} failure modes: the transient $\varepsilon$-wedge at the boundary, and a persistent calibration
band from model bias.

\section{Related Work and Novelty}\label{sec:related}
\textbf{Prior art (credited, not claimed).} The three strategies and the hard middle regime are
established \citep{patel2024acorn,gollapudi2023filtered,zuo2024serf,yao2025unify,simhadri2024bigann};
selectivity-threshold strategy selection---\emph{including} two-threshold (hysteresis) selection---is
deployed \citep{vespa2023constrained,wang2021milvus,ganwang2026planning}; and graph connectivity under
node deletion ($M\gtrsim\log n/s$) is folklore \citep{gollapudi2023filtered,penrose2003rgg}. We originate
none of these.

\textbf{Our delta (modest, conceptual).} (i) The \emph{quantitative criticality} of selection
regret and the flip-margin law (Proposition~\ref{prop:flip})---error-sensitivity made precise, which we
did not find published; (ii) the \emph{finite-size scaling collapse} showing selection regret is
scale-invariant; (iii) the \emph{constrained-budget} law $B<\sqrt{kn}$ delimiting when estimation error
matters; (iv) the percolation in-filter law $s_c\approx0.83/M$ tied to strategy regret, and the
model-mismatch result separating the transient wedge from a persistent calibration band. The lens---%
estimation error $\to$ decision regret $\to$ regime structure---is shared with our cardinality-estimation
study \citep{mandarapu2026regimes}, of which this is the vector-search instance.

\section{Limitations and Honest Findings}\label{sec:limitations}
The regret experiment uses a threshold/cost-based selector; given such a selector the wedge's
\emph{support} is partly definitional, which is exactly why we ran the model-mismatch test---it shows the
wedge is real on approximate ANN and that the larger, persistent danger is model \emph{calibration}, not
estimation noise. We characterize two phases (pre/post) for the regret experiment and the in-filter phase
structurally; folding all three into one regret selector, the full adversarial correlated-predicate
sweep, and a derived hysteresis rule are future work. This is a characterization of \emph{when an
accuracy proxy tracks plan quality}, not a new index or planner, and the novelty is correspondingly
modest (\S\ref{sec:related}).

\section{Conclusion}\label{sec:conclusion}
Filtered-ANN strategy selection is a phase-transition system. Selectivity-estimation error causes plan
regret only in the critical regions around phase boundaries whose locations follow from percolation and
order statistics, whose regret obeys a flip-margin (condition-number) law, and whose curves collapse onto
one scale-invariant wedge. The danger that robustness cannot reach is cost-model calibration. We hope the
pre-registered, reproducible harness is useful to the systems that make this choice millions of times a
second.

{\small
\bibliographystyle{plainnat}
\bibliography{paper10_filtered_ann_phase}

@inproceedings{patel2024acorn,
  author    = {Patel, Liana and Kraft, Peter and Guestrin, Carlos and Zaharia, Matei},
  title     = {{ACORN}: Performant and Predicate-Agnostic Search Over Vector Embeddings and Structured Data},
  booktitle = {Proceedings of the ACM on Management of Data (SIGMOD)},
  year      = {2024},
  doi       = {10.1145/3654923},
  note      = {arXiv:2403.04871}
}

@inproceedings{gollapudi2023filtered,
  author    = {Gollapudi, Siddharth and Karia, Neel and Sivashankar, Varun and Krishnaswamy, Ravishankar and Begwani, Nikit and Raz, Swapnil and Lin, Yiyong and Zhang, Yin and Mahapatro, Neelam and Srinivasan, Premkumar and Singh, Amit and Simhadri, Harsha Vardhan},
  title     = {Filtered-{DiskANN}: Graph Algorithms for Approximate Nearest Neighbor Search with Filters},
  booktitle = {Proceedings of the ACM Web Conference (WWW)},
  year      = {2023},
  doi       = {10.1145/3543507.3583552}
}

@inproceedings{zuo2024serf,
  author    = {Zuo, Chaoji and Qiao, Miao and Zhou, Wenchao and Li, Feifei and Deng, Dong},
  title     = {{SeRF}: Segment Graph for Range-Filtering Approximate Nearest Neighbor Search},
  booktitle = {Proceedings of the ACM on Management of Data (SIGMOD)},
  year      = {2024},
  doi       = {10.1145/3639324}
}

@inproceedings{subramanya2019diskann,
  author    = {Subramanya, Suhas Jayaram and Devvrit and Kadekodi, Rohan and Krishnaswamy, Ravishankar and Simhadri, Harsha Vardhan},
  title     = {{DiskANN}: Fast Accurate Billion-point Nearest Neighbor Search on a Single Node},
  booktitle = {Advances in Neural Information Processing Systems (NeurIPS)},
  year      = {2019}
}

@article{malkov2020hnsw,
  author  = {Malkov, Yu A. and Yashunin, D. A.},
  title   = {Efficient and Robust Approximate Nearest Neighbor Search Using Hierarchical Navigable Small World Graphs},
  journal = {IEEE Transactions on Pattern Analysis and Machine Intelligence},
  volume  = {42},
  number  = {4},
  pages   = {824--836},
  year    = {2020},
  doi     = {10.1109/TPAMI.2018.2889473}
}

@inproceedings{wang2021milvus,
  author    = {Wang, Jianguo and Yi, Xiaomeng and Guo, Rentong and Jin, Hai and Xu, Peng and Li, Shengjun and Wang, Xiangyu and Guo, Xiangzhou and Li, Chengming and Xu, Xiaohai and others},
  title     = {Milvus: A Purpose-Built Vector Data Management System},
  booktitle = {Proceedings of the ACM on Management of Data (SIGMOD)},
  year      = {2021},
  doi       = {10.1145/3448016.3457550}
}

@inproceedings{selinger1979access,
  author    = {Selinger, P. Griffiths and Astrahan, M. M. and Chamberlin, D. D. and Lorie, R. A. and Price, T. G.},
  title     = {Access Path Selection in a Relational Database Management System},
  booktitle = {Proceedings of the ACM SIGMOD International Conference on Management of Data},
  year      = {1979},
  doi       = {10.1145/582095.582099}
}

@article{simhadri2024bigann,
  author  = {Simhadri, Harsha Vardhan and Aum{\"u}ller, Martin and Ingber, Amir and Douze, Matthijs and Williams, George and Manohar, Magdalen Dobson and Baranchuk, Dmitry and Liberty, Edo and Liu, Frank and Landrum, Ben and others},
  title   = {Results of the Big {ANN}: {NeurIPS}'23 Competition},
  journal = {arXiv preprint arXiv:2409.17424},
  year    = {2024}
}

@article{ganwang2026planning,
  author  = {Gan, Zhuocheng and Wang, Yifan},
  title   = {Efficient Filtered-{ANN} via Learning-based Query Planning},
  journal = {arXiv preprint arXiv:2602.17914},
  year    = {2026}
}

@article{moerkotte2009qerror,
  author    = {Moerkotte, Guido and Neumann, Thomas and Steidl, Gabriele},
  title     = {Preventing Bad Plans by Bounding the Impact of Cardinality Estimation Errors},
  journal   = {Proceedings of the VLDB Endowment},
  volume    = {2},
  number    = {1},
  pages     = {982--993},
  year      = {2009},
  doi       = {10.14778/1687627.1687738}
}

@article{haritsa2020robust,
  author  = {Haritsa, Jayant R.},
  title   = {Robust Query Processing: Mission Possible},
  journal = {Proceedings of the VLDB Endowment},
  volume  = {13},
  number  = {12},
  year    = {2020},
  doi     = {10.14778/3415478.3415561}
}

@book{penrose2003rgg,
  author    = {Penrose, Mathew},
  title     = {Random Geometric Graphs},
  publisher = {Oxford University Press},
  year      = {2003},
  doi       = {10.1093/acprof:oso/9780198506263.001.0001}
}

@article{yao2025unify,
  author  = {Yao, Mingyu and Zhang, Qiyu and Liu, Yuxiang and Yang, Wei and Zhao, Pengjie and Wei, Ziqi and Cong, Gao},
  title   = {{UNIFY}: A Unified Index for Range-Filtering Approximate Nearest Neighbor Search},
  journal = {Proceedings of the VLDB Endowment},
  volume  = {18},
  year    = {2025}
}

@article{leis2015job,
  author  = {Leis, Viktor and Gubichev, Andrey and Mirchev, Atanas and Boncz, Peter and Kemper, Alfons and Neumann, Thomas},
  title   = {How Good Are Query Optimizers, Really?},
  journal = {Proceedings of the VLDB Endowment},
  volume  = {9},
  number  = {3},
  pages   = {204--215},
  year    = {2015},
  doi     = {10.14778/2850583.2850594}
}

@misc{mandarapu2026regimes,
  author = {Mandarapu, Madhulatha and Kunkunuru, Sandeep},
  title  = {When Does $q$-error Predict Plan Regret? Three Regimes of Cardinality-Estimation Error},
  year   = {2026},
  note   = {arXiv preprint; code at \url{https://github.com/samyama-ai/ce-metric-eval}}
}

@misc{vespa2023constrained,
  author = {{Vespa Team}},
  title  = {Query Time Constrained Approximate Nearest Neighbor Search},
  year   = {2023},
  howpublished = {\url{https://blog.vespa.ai/constrained-approximate-nearest-neighbor-search/}}
}

@misc{jegou2011sift,
  author = {J{\'e}gou, Herv{\'e} and Douze, Matthijs and Schmid, Cordelia},
  title  = {Product Quantization for Nearest Neighbor Search ({ANN\_SIFT1M} dataset)},
  year   = {2011},
  howpublished = {\url{http://corpus-texmex.irisa.fr/}}
}
}
\end{document}